\documentclass[sigconf]{acmart}
\usepackage{hyperref}
\usepackage{hyperxmp}
 
\usepackage[caption=false]{subfig}

\copyrightyear{2023}
\acmYear{2023}
\setcopyright{acmlicensed}

\acmConference[HAI '23] {International Conference on Human-Agent Interaction}{December 4--7, 2023}{Gothenburg, Sweden}

\acmBooktitle{International Conference on Human-Agent Interaction (HAI '23), {December 4--7, 2023, Gothenburg, Sweden}}

\acmPrice{15.00}

\acmDOI{10.1145/3623809.3623929}

\acmISBN{979-8-4007-0824-4/23/12}

\begin{document}
\title{How do Humans take an Object from a Robot: \\Behavior changes observed in a User Study}

\author{Parag Khanna}
\affiliation{%
  \institution{KTH Royal Institute of Technology}
  \city{Stockholm}
  \country{Sweden}}
\email{paragk@kth.se}

\author{Elmira Yadollahi}
\affiliation{%
  \institution{KTH Royal Institute of Technology}
  \city{Stockholm}
  \country{Sweden}}
\email{elmiray@kth.se}

\author{Iolanda Leite}
\affiliation{%
  \institution{KTH Royal Institute of Technology}
  \city{Stockholm}
  \country{Sweden}}
\email{iolanda@kth.se}

\author{Mårten Björkman}
\affiliation{%
  \institution{KTH Royal Institute of Technology}
  \city{Stockholm}
  \country{Sweden}}
\email{celle@kth.se}

\author{Christian Smith}
\affiliation{%
  \institution{KTH Royal Institute of Technology}
  \city{Stockholm}
  \country{Sweden}}
\email{ccs@kth.se}
\renewcommand{\shortauthors}{Khanna et al.}
\begin{abstract}
To facilitate human-robot interaction and gain human trust, a robot should recognize and adapt to changes in human behavior. This work documents different human behaviors observed while taking objects from an interactive robot in an experimental study,
categorized across two dimensions: pull force applied and handedness. We also present the changes observed in human behavior upon repeated interaction with the robot to take various objects.
\end{abstract}
\maketitle
\section{Introduction}
A robot must be able to read subtle changes in human behavior and respond accordingly, similar to humans, in order to improve human-robot interaction and garner human trust \cite{review_adapt_robot_behavior,robot_adaptation_trust}.
In this work, we discuss different human behaviors observed while taking objects from a robot in robot-to-human handovers. Our study tested a popular robotic grip release technique for various objects in a repeated handover scenario with novice users. Further, we highlight the changes observed in the behavior of novice users in repeated handovers with the robot within the same interactive experiment. 
\vspace{-1.15mm}
\section{Experiments}
\subsection{Procedure}
This experimental study is described in detail in \cite{khanna2023user,exp_strategies_roman_khanna2023}
and involve a collaborative task between an interactive robot and a human with the goal of filling a shelf with objects (Fig. 1a).
The participants had no prior experience of physical interaction with the robot and were given no prior information about the robot's abilities other than that it was interactive and fully autonomous.
The experiment involved 4 rounds of 4 objects each, where the human would place these objects in front of the robot to start the task.
The robot had to execute \textit{Pick}, \textit{Carry}, and \textit{Place} actions for each object to successfully transfer that object to the shelf. We incorporated pre-programmed robotic failures for each action based on the type of object without the participant's knowledge. At each failure, the robot explains the failure and asks for human help as part of the resolution provided to the human. Each participant would undergo the same order of objects and robotic failures while being exposed to different explanation levels at each round \cite{exp_strategies_roman_khanna2023}.
This work analyzes the data from  68 participants that participated in the experiment.
\begin{figure}[t]
      \centering
         \subfloat[]{
     \includegraphics[width=4.3cm,trim={5.5cm 2.4cm 13cm 5cm},clip]{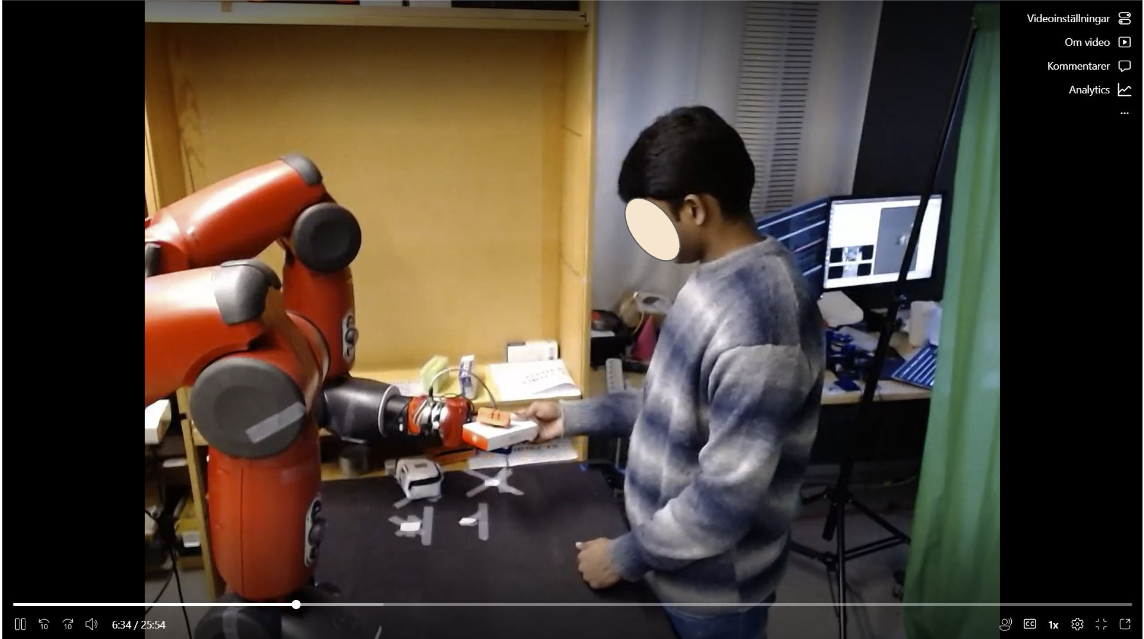}}
     \subfloat[]{
     \includegraphics[width=4.01cm,height=3.0cm,trim={0cm 4cm 14cm 0cm},clip]{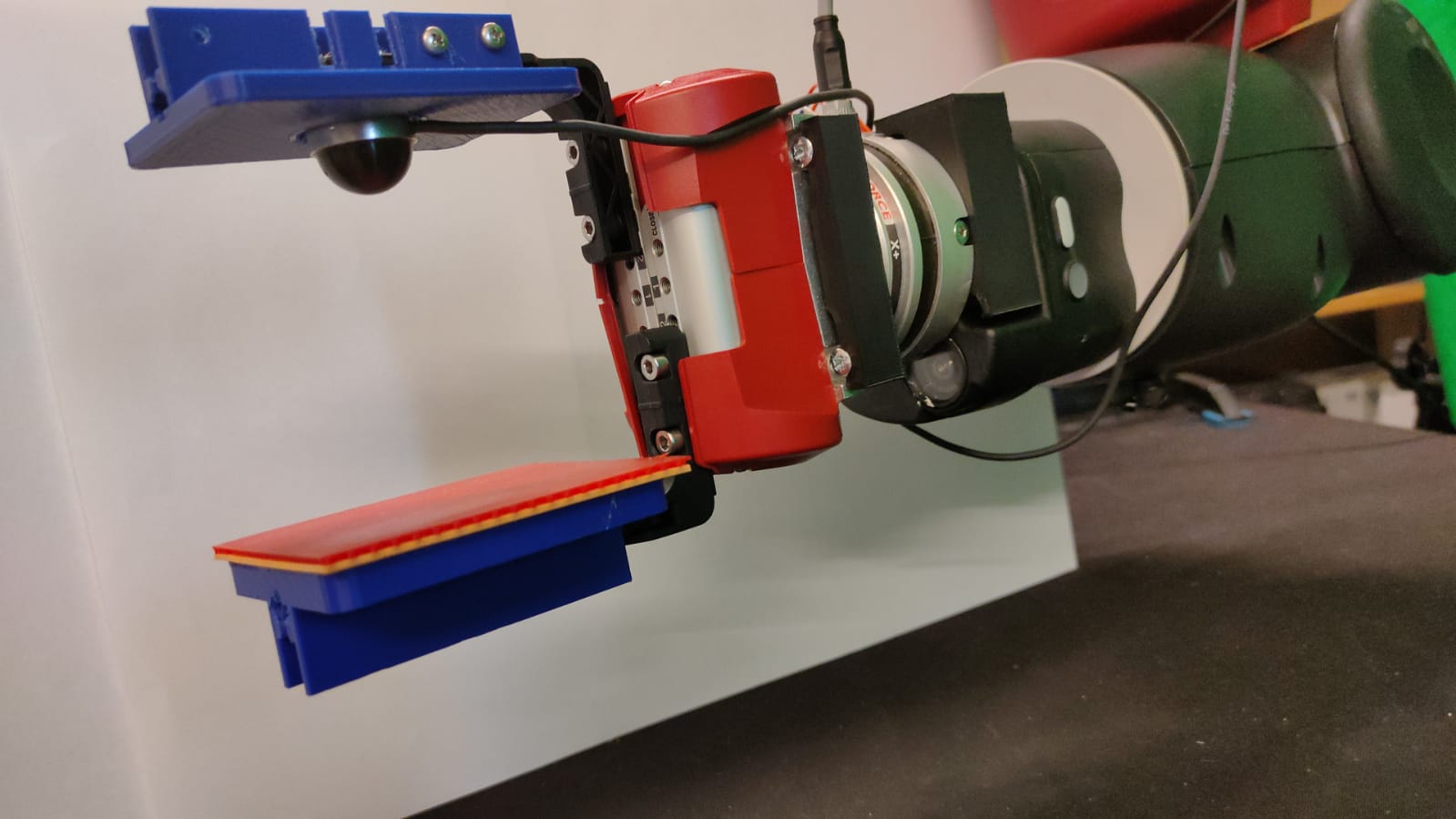}}  
    \setlength\abovecaptionskip{-0.001\baselineskip}
    
      \caption{(a) A snippet of a robot-to-handover in the study, (b) Gripper with a big base mounted on robot's wrist sensor}
      \label{fig:H2H_handover_study}
\end{figure}
\subsection{Robot-to-Human Handovers}
In the study, there was a repeated occurrence of human-robot handovers, which was incorporated as part of the resolutions after failures so the human and robot collaboratively complete the task.
The robot required human help in the form of human-to-robot handovers for the objects it could not pick. On the other hand, robot-to-human handover occurred when the robot failed to carry the object due to excessive weight or if it failed to place the object at the desired location on the shelf. These robot-to-human handovers were followed by the human carrying and placing the object on the shelf, thus completing the task.

In this work, we focus on the robot-to-human handovers that formed the resolution action for \textit{Carry} and \textit{Pick} robotic failures; and were needed for 8 objects in each experiment per participant. This handover occurred after the robot moved to a fixed handover pose, carrying the object horizontally (Fig. 1a), and said "Can you please take the object back?".

\emph{\textbf{Grip Release}}: For the automatic opening of the gripper by the robot to release the object, a pull force thresholding-based grip release strategy was adopted, which is a popular technique in the literature \cite{pull_proactive_strategy_8673085,loadsharing_pull_strategy-10.3389/frobt.2021.672995,chan_grip_from_load_second_PR2,parag-humanoids}. 
The robot would release the object if the human applied a minimal pull force of 3N. This pull force was measured along the direction of handover by a Force/Torque sensor mounted at the wrist of the robot \cite{parag-humanoids}.

\emph{Timed Automatic Grip-Release}: The robot was also programmed for an automatic grip release if no or insufficient pull was observed until 10 seconds after the robot's request to take the object back. In this case, the object did not fall out of the gripper due to the large gripper base (Fig. 1b), and the human could simply take the object out of the gripper.
\section{Human Behavior in handovers}
We consider 2 dimensions across which we analyzed human behaviors: \emph{Pulling} and \emph{Handedness}.
\subsection{Pulling Behavior}
\begin{table}[b]
\setlength\abovecaptionskip{-0.05\baselineskip}
\begin{center}
\caption{Description of changes in Behavior}
\begin{tabular}{|c|c|c|c|} \toprule
       \textbf{$\Delta_B$} & \textbf{Magnitude} & \textbf{$\Delta$Pull Behavior} & \textbf{$\Delta$Handedness} \\\midrule
     3 & Large & PF or PS $\leftrightarrow$ HNP & YES or NO \\
     2 & Moderate & PF $\leftrightarrow$ PS & YES or NO \\
     1 & Small & NO & YES \\
     0 & None & NO & NO \\\bottomrule
\end{tabular}
\end{center}
\end{table}
We observed different human behaviors while taking the object from the robot and classified them into 3 categories:
\subsubsection{Pull Fine (PF)} The robotic grip release occurred as the human held and pulled the object out, removing the object from the gripper within 3 seconds of first human contact with the object. This was expected behavior in line with a prior study with this robotic platform and the pilot study with experienced users. 
\subsubsection{Pull Slow (PS)} 
It takes longer than 3 seconds for the human to apply sufficient pull to take the object after their first hold or touch on it. Upon holding the object, the human applies a little to no pull as they figure out how the robot releases the object. However, they increase the applied pulling force once they firmly held the object and observed that the robot had not opened its grip. The robot finally opens the gripper as a sufficient pull force is measured before the timed automatic grip release is triggered. 
This behavior shows the human taking object from the robot with a little caution.

\subsubsection{Hold, no pull with Verbal Commands (HNP)}
The human just holds on to the object without a sufficient pull or any pull force at all. In this scenario, some participants also try to give verbal commands or requests to the robot to release the object. This shows the human understanding of robots as completely verbal and expects the robot to follow their commands. However, the robot only releases the object via timed automatic grip release, i.e., 10 seconds after the robot's request to take the object.
This behavior represents an extremely cautious approach in taking the object from the robot, as the human waited for the gripper to open before taking the object.

\subsection{Handedness} We further propose that an additional dimension to analyze human behavior can be inspected, if they used one hand or two hands to take the object. We observed that some participants used two hands while taking the object back indicating extra cautious approach.   

\section{Changes in Human Behavior}
For several participants, we were able to observe a change in their handover behavior in repeated handovers, as they took the object from the robot again. We propose a measure of change in behavior as $\Delta_B$ whose magnitude increases with increased change in behavior. Table 1 describes the proposed values for $\Delta_B$ in accordance with different behaviors. The priority is given to a change in pull behavior over handedness. A large $\Delta_B$ corresponds to a participant changing behavior from pulling fine/pulling slowly to just holding the object while not pulling at all or vice versa. A moderate $\Delta_B$ corresponds to a change from pulling fine to pulling slow, or vice versa. If there is only a change in handedness, i.e., a person changed from using one hand to two hands or vice versa, the $\Delta_B$ is assigned a value of 1. A $\Delta_B$ of 0 corresponds to no change observed.
\begin{table}[t]
\setlength\abovecaptionskip{-0.05\baselineskip}
\begin{center}
\caption{Number of participants and $\Delta_B$}
\begin{tabular}{|c|c|c|c|c|} \toprule
       \textbf{$\Delta_B$} & \textbf{0} & \textbf{1} & \textbf{2} & \textbf{3}\\\midrule
       Number & 20 & 4 & 17& 27 \\\midrule
       Sum & 20 & \multicolumn{3}{c|}{48} \\ 
       \bottomrule
\end{tabular}
\end{center}
\end{table}

Further, Table 2 displays the number of participants for whom a specific level of $\Delta_B$ was observed. Out of 68, only 20 participants showed no change in behavior, while 48 showed changes in behavior. 
Moreover, there is a different magnitude of behavior changes observed for different participants, with a high number of people showing a large (27) and moderate (17) change, respectively. 
\section{Conclusion and Future Work}
The occurrence of different pull force-based human behaviors in handovers make it necessary for a robot to plan and adapt its handover strategies. Rather than relying on a fixed grip release strategy like pull force-based thresholding, the strategy should be personalized according to the current user and consider other modalities such as verbal communication

However, we also observe that there can be changes in human handover behavior with repeated handovers with the robot. This shows humans adapt to the robot based on the interaction.
Accordingly, the robot should be able to modify its handover strategy online during the same interaction in depending on user behavior.
This leads us to a challenging future work that involves studying the factors leading to the changes in human handover behavior. We would further investigate these factors by designing an experiment solely aimed at identifying these behaviors and aim to impart a robot with the capability to observe these factors online and adapt its strategies in advance for a better user experience.

\bibliographystyle{ACM-Reference-Format}
\bibliography{pk_ref2_updated}

\end{document}